\documentclass[lettersize,journal]{IEEEtran}
\usepackage{amsmath,amsfonts}
\usepackage{algorithmic}
\usepackage{algorithm}
\usepackage{array}
\usepackage[caption=false,font=normalsize,labelfont=sf,textfont=sf]{subfig}
\usepackage{textcomp}
\usepackage{stfloats}
\usepackage{url}
\usepackage{verbatim}
\usepackage{graphicx}
\usepackage{cite}
\usepackage{booktabs}
\usepackage{algorithm}
\usepackage{amsmath}
\usepackage{amssymb}
\usepackage{mathtools}
\usepackage{amsthm}
\usepackage{multirow}
\usepackage{color,lineno,hyperref}
\usepackage{amsthm,amsmath,amssymb}
\usepackage{mathrsfs}

\hypersetup{colorlinks = true, 
	    linkcolor = black, 
	    urlcolor = blue,
           citecolor = blue} 
           
\begin{document}

\title{In-Distribution Consistency Regularization Improves the Generalization of Quantization-Aware Training}
\author{
Junbiao Pang, Tianyang Cai, Jiaqi Wu, Baochang Zhang

\IEEEcompsocitemizethanks{
\IEEEcompsocthanksitem J. Pang, T. Cai and J. Wu are with the Faculty of Information Technology, Beijing University of Technology, Beijing 100124, China (e-mail: \mbox{junbiao\_pang@bjut.edu.cn}).

\IEEEcompsocthanksitem  B. Zhang is with the University of Chinese Academy of Sciences, Chinese Academy of Sciences (CAS), Beijing
100049, China, and the Institute of Computing Technology, CAS, Beijing
100190, China (email: bczhang@139.com).
 }

}

\maketitle

\begin{abstract}


Although existing Quantization-Aware Training (QAT) methods intensively depend on knowledge distillation to guarantee performance, QAT still suffers from severe performance drop. The experiments have shown that vanilla quantization is sensitive to the perturbation from both the input and weights. Therefore, we assume that the generalization ability of QAT is predominantly caused by both the intrinsic instability (training time) and the limited generalization ability (testing time). In this paper, we address both issues from a new perspective by leveraging Consistency Regularization (CR) to improve the generalization ability of QAT. Empirical results and theoretical analysis verify that CR would bring a good generalization ability to different network architectures and various QAT methods. Extensive experiments demonstrate that our approach significantly outperforms current state-of-the-art QAT methods and even the FP counterparts. On CIFAR-10, the proposed method improves by 3.79\% compared to the baseline method using ResNet18, and improves by 3.84\% compared to the baseline method using the lightweight model MobileNet. 

\end{abstract}


\section{Introduction}

Model compression has emerged as an inevitable imperative to deploy deep models on edge devices. Classical approaches for the model compression include neural architecture search~\cite{xiao-nas-cvpr2022}, network pruning~\cite{bai-sparsellm-nips-2024} and quantization~\cite{jacob-quantization-cvpr-2018}\cite{xu-qdetr-cvpr-2023}\cite{Li-quant1-tnnls-2024}. Quantization depends on low-precision arithmetic, resulting in lower latency, smaller memory footprint, and less energy consumption. Quantization has two research lines: Post-Training Quantization (PTQ)~\cite{nagel-adaround-pmlr-2020}\cite{li-brecq-arxiv-2021}\cite{wei-qdrop-arxiv-2022}\cite{Li-PSAQ-tnnls-2023} and Quantization-Aware Training (QAT)~\cite{esser-lsq-arxiv-2019}\cite{nagel-ooq-arxiv-2022}\cite{Wang-tab-tnnls-2024}. In particular, QAT try to preserve the accuracy of NNs in a low-bit-width model by retraining NNs~\cite{esser-lsq-arxiv-2019}\cite{nagel-ooq-arxiv-2022} or utilizing Knowledge Distillation (KD)~\cite{wei-qkd-eccv-2018}\cite{li-qvit-nips-2022}. However, the accuracy of QAT is still lower than that of its FP counterpart at low bits. However, QAT uses complete datasets and has expensive computational overhead. Naturally, one question is what makes the Quantized NNs have inferior performance.

To mitigate the gap, previous research on Low-Bit Quantization networks (LBQ) has focused on designing more effective optimization algorithms to find better local minima of quantized weights~\cite{li-brecq-arxiv-2021}~\cite{wei-qdrop-arxiv-2022}. However, the task is highly nontrivial, since the optimization that used to be effective to train
Deep Neural Networks (DNNs) now becomes tricky to implement, \textit{e.g.}, formulating round operation as optimization problem~\cite{nagel-adaround-pmlr-2020}, adding the scale and the zero point as parameters~\cite{esser-lsq-arxiv-2019}\cite{bhalgat-lsq+-cvpr-2020}

In this paper, we investigate LBQ systematically in terms of robustness and generalization ability. We find that LBQ suffers from severe intrinsic non-robustness and low generalization ability.
The observation implied that the performance degradation of LBQ is not likely to be
resolved by solely improving the optimization techniques~\cite{esser-lsq-arxiv-2019}\cite{bhalgat-lsq+-cvpr-2020}; instead, it is mandatory to improve the robustness to noises and boost its generalization ability.

\begin{figure}[t!]
    \centering
    \includegraphics[width=1.0\linewidth]{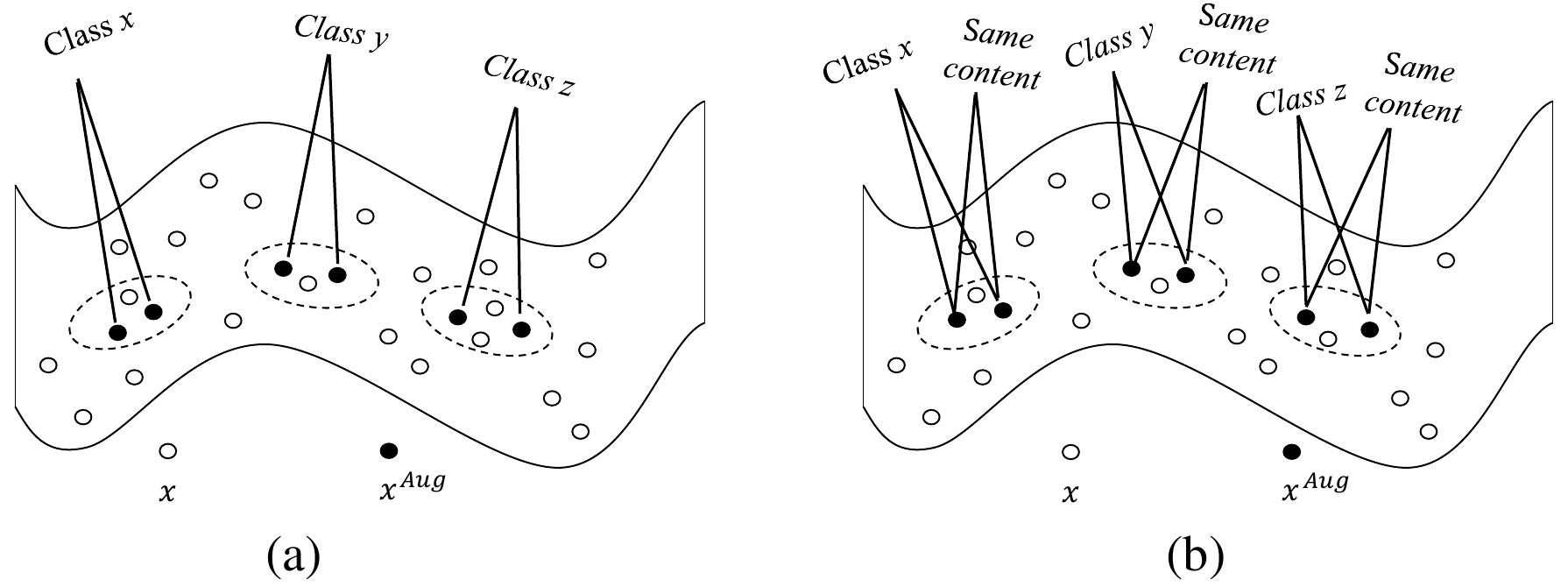}
    \caption{Comparison between KD and CR, where $x$ represents a origin sample, $x^{Aug}$ represents the augmented sample. (a) KD is used on individual augmented samples. (b) CR models the consistency between two augmented samples.}
    \label{fig:manifold}
\end{figure}

Inspired by the analysis, in this work, we propose to exploit the robustness and generalization ability from in-distribution dataset (including unlabeled samples). The basic assumption of in-distribution is from a sampled should invariance to any perturbation~\cite{moosavi-perturbations-cvpr-2017}, augmentation~\cite{zhang-mixup-arxiv-2017}~\cite{devries-cutout-arxiv-2017} or attacking~\cite{zhao-attack-cvpr-2020}. A naive approach is to increase the types of data augmentation or noises for the conventional QAT methods in Fig.~\ref{fig:manifold}(a). However, this simple approach assumes that an augmented images belong to the same class, which is extensively adopted in the traditional approaches~\cite{esser-lsq-arxiv-2019,bhalgat-lsq+-cvpr-2020}. 

In this paper, to inject generalization ability into quantized NNs, we exploit the fact that two augmented images still describe the same content in Fig.~\ref{fig:manifold} (b). That is, we intuitively expects a quantized network stable under the \textit{any} change of the same samples. We show that the statistical properties of the quantized NNs become much nicer: the robustness to noises and the generalization ability of quantized models at test time are significantly improved. Moreover, we borrow the concept of  sharpness~\cite{keskar-eigenvalue1-iclr-2017} from statistical learning theory to describe the generalization ability of deep models. We theoretically show that the minimizing consistency among augmented images is equal to minimize the sharpness of loss landscape.

\begin{figure}[t!]
    \centering
    \includegraphics[width=1.0\linewidth]{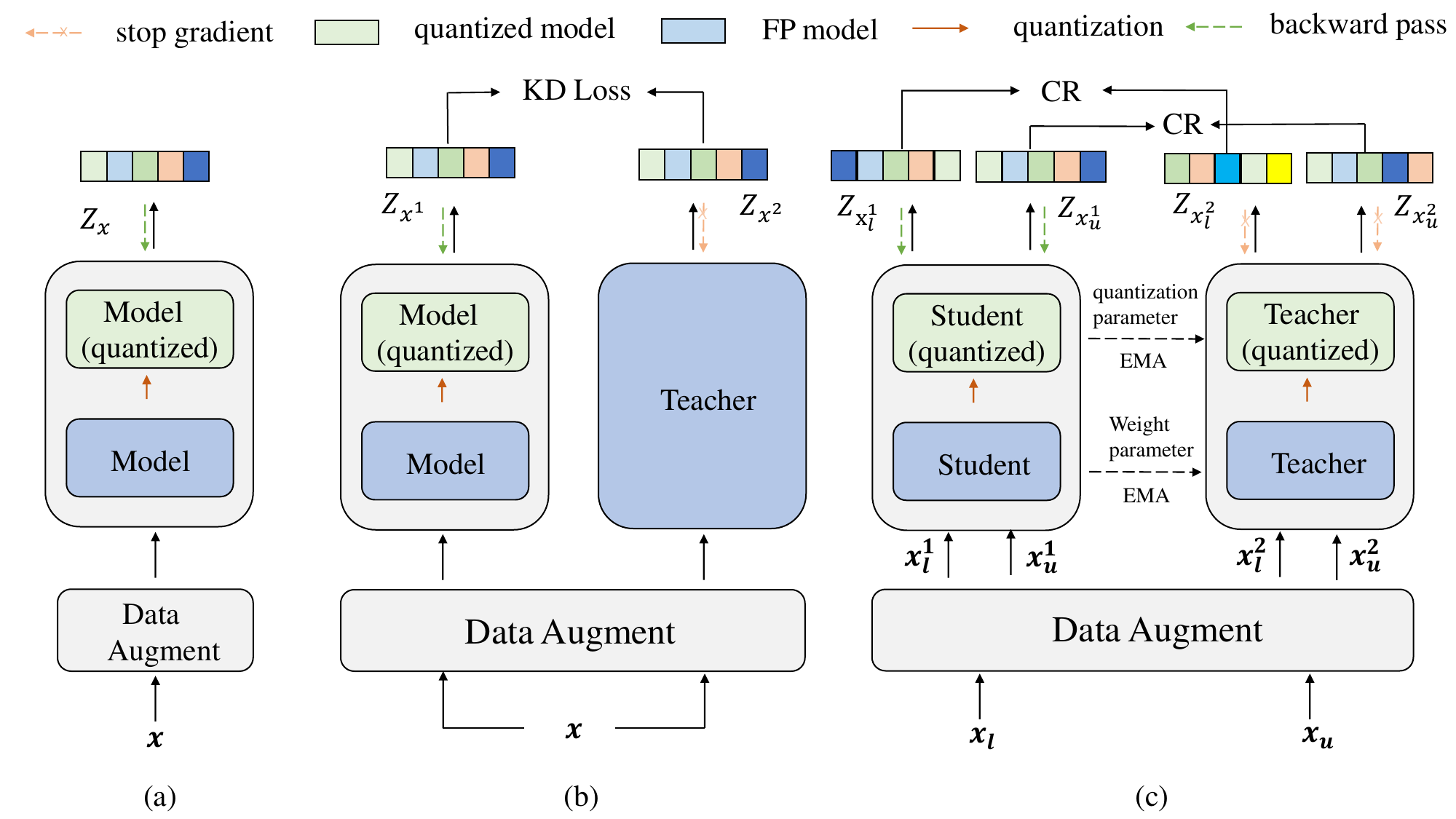}
    \caption{Comparison between vanilla QAT (a), quantization with KD~\cite{polino-qkd-arxiv-2018}(b), and our method (c).}
    \label{fig:qat-kd-cr-comparision}
\end{figure}

In this paper, we propose Consistency Regularization (CR) to convert the perturbation between two augmented samples into the generalization ability of quantized NNs. Rather than adopting a standard data arguments (which is also the most used pre-processing method) in Fig.~\ref{fig:qat-kd-cr-comparision} (a) and KD~\cite{li-qvit-nips-2022}~\cite{xu-idanet-eccv-2022} (which leverages the logits as a teacher) in Fig.~\ref{fig:qat-kd-cr-comparision} (b), the proposed CR mines consistency to retain the generalization ability in Fig.~\ref{fig:qat-kd-cr-comparision} (c). Concretely, a student network represents \emph{any} QAT model, while a teacher network is a temporal ensemble of student networks in the parameter space. 
We summarize our contributions as follows:

\begin{itemize}
\item To our best knowledge, we utilize the sharpness from statistical learning theory to boost the generalization ability of QAT. We theoretically verify that minimizing CR between two augmented samples would minimize the sharpness of the quantized network, if the multiplication of Lipschitz constant for each layer is not approximately to zero.
\item We show that our method could uniformly leverage the supervision signals either from labeled samples or the in-distribution unlabeled ones. The proposed approach, a simple, novel, yet powerful method, is easily adapted to different neural networks. CR pushes the performances of QAT models towards, and even surpasses that of FP32 counterparts.
\item Extensive experiments on different models prove that our method set up a new SOTA method for QAT, validating the power of unlabeled data in model quantization. 
On CIFAR-10, the proposed method improves by 3.79\% compared to the baseline method using ResNet18, and improves by 3.84\% compared to the baseline method using the lightweight model MobileNetV1. On CIFAR-100, the proposed method improves by 0.71\% compared to the baseline method using ResNet50, and improves by 2.17\% compared to the baseline method using the lightweight model MobileNetV2.
\end{itemize}

\section{Related Work}

\textbf{Quantization-Aware Training.} QAT minimizes quantization errors by re-training with the task-related loss on the whole dataset. There are two research directions in QAT: 1) one primarily focuses on addressing quantization-specific problem~\cite{bengio-ste-arxiv-2013}~\cite{lee-ewgs-cvpr-2021}~\cite{nagel-ooq-arxiv-2022}; and 2) the other focuses on the other aspects of quantization, \textit{e.g.}, robust to adversarial noise~\cite{lin-defensive-iclr-2019}, reduce variance of prediction~\cite{zhu-benn-cvpr-2019}.

For the first category, most researches focus on specific problems for some network architecture, \textit{e.g.}, gradient bias and weight oscillations in QAT training. For instance, Straight-Through Estimator (STE)~\cite{bengio-ste-arxiv-2013} used the expected probability of stochastic quantization as the gradient for back-propagation. Element-Wise Gradient Scaling (EWGS)~\cite{lee-ewgs-cvpr-2021} adaptively scaled quantized gradients based on quantization error. Position-based Scaled Gradient (PSG)~\cite{kim-psg-nips-2020} scaled gradients based on the position of the weight vector to handle gradient compensation problem. Differentiable Quantization (DiffQ)~\cite{defossez-diffq-arxiv-2021} reduced the bias of STE with the simulated quantization noises. Oscillations Quantization (OOQ)~\cite{nagel-ooq-arxiv-2022} addressed oscillation issues by encouraging weights to be as close to the center of the bin as possible. 

For the second category, some importance aspects of QAT, \textit{e.g.}, robust~\cite{chmiel-robustquantization-nips-2020}, adaptive bit-widths~\cite{jin-adabits-cvpr-2020}, and Network Architecture Search (NAS)~\cite{shen-once-cvpr-2021}, are researched for the QAT training paradigms. For instance, Synergistic Self-supervised and Quantization Learning (SSQL)~\cite{cao-ssql-eccv-2022} proposed to learn quantization-friendly representations.~\cite{Wang-quant2-tnnls-2021} combined unsupervised learning and binary quantization to maintain model performance. \cite{wei-qkd-eccv-2018} effectively combined quantization and distillation to induce light-weight networks.
To our best knowledge, our method bridges the generalization from the statical learning with QAT to improve the robustness and generalization ability.  

\textbf{Data Augmentation}. Recently, several studies have demonstrated that data augmentation could enhance the performances. For instance, \cite{gulrajani-search-arxiv-2020} found that data augmentation could outperform various domain generalization methods. \cite{zhang-mixup-arxiv-2017} introduced Mixup that combines the sample pairs linearly to generate a new sample and label. Cutout~\cite{devries-cutout-arxiv-2017} randomly removed one or more rectangular areas to generate diverse samples. \cite{ilse-selecting-icml-2021} contributed to select appropriate data augmentation strategies. In this paper, we utilize data augmentation to generate reasonable perturbation. 

\textbf{Injecting Generalization for Quantization.} The generalization ability of quantized models is an interesting problem. For instance, a very recent work considered a sharpness-aware approach and
evaluated its performance from noisy labels or data~\cite{li-improved-nips-2021}. Stochastic Weight Averaging (SWA)~\cite{yang-swa-icml-2019} averaged the end of several periodic for a flatness minima. QDrop~\cite{wei-qdrop-arxiv-2022} introduced a noise scheme for a flatness of loss landscape. However, their approach is not readily adopted by QAT.
Different from the perturbation of weights is SAM, the perturbation from images is used in our method. We defer a more extensive discussion two approaches in Section~\ref{sec:reason}.


\section{Methodology}


\subsection{Notation and Background}


\textbf{Basic Notations.} In this paper, notation $\boldsymbol{x}$ represents a matrix (or tensor), a vector is denoted as  $\boldsymbol{x}$, $f(\boldsymbol{x};\boldsymbol{w}$) represents a FP model with the weight parameter $\boldsymbol{w}$ and the input $\boldsymbol{x}$, $f(\boldsymbol{x};\boldsymbol{w}, \boldsymbol{s}, \boldsymbol{z})$ represents a quantized model with the parameter $\boldsymbol{w}$, quantization parameter $\boldsymbol{s}$, $\boldsymbol{z}$ and the input $\boldsymbol{x}$. We assume sample $\boldsymbol{x}$ is generated from the training set $\mathscr{D}_{t}$. 

\textbf{Quantization.} Step size $\boldsymbol{s}$ and zero points $\boldsymbol{z}$ serve as a bridge between floating-point and fixed-point representations. Given the input tensor $\boldsymbol{x}$\footnote{It could either be feature $\boldsymbol{x}$ or weight $\boldsymbol{w}$.}, the quantization operation is as follows:
\begin{equation}\label{eqt:quantization}
    \begin{aligned}
    \boldsymbol{x}_{int} &= clip\left (   \lfloor{\frac{\boldsymbol{x}}{\boldsymbol{s}}}\rceil + \boldsymbol{z},0,2^{q}-1 \right ),\\
    \hat{\boldsymbol{x}} &=\left ( \boldsymbol{x}_{int}-\boldsymbol{z}  \right ) \boldsymbol{s},
    \end{aligned}
\end{equation}
where $\lfloor{\cdot  }\rceil$ represents the rounding-to-nearest operator, $q$ is the predefined quantization bit-width, $\boldsymbol{s}$ denotes the scale between two subsequent quantization levels.  $\boldsymbol{z}$ stands for the zero-points. Both $\boldsymbol{s}$ and $\boldsymbol{z}$ are initialized by a calibration set $\mathscr{D}_{c}$ from the training dataset $ \mathscr{D}_{t}$, \textit{i.e.}, $\mathscr{D}_{c}\in \mathscr{D}_{t}$.
\begin{eqnarray}
\boldsymbol{s} =\frac{\boldsymbol{x}_{max} - \boldsymbol{x}_{min}}{2^{q} - 1},\label{eqt:scale_init} \\
\boldsymbol{z} =\lfloor{q_{max}-\frac{\boldsymbol{x}_{max}}{\boldsymbol{s}}}\rceil,\label{eqt:zero_init}
\end{eqnarray}
where $q_{max}$ is the maximum value of the quantized integer.

We follow the practice in LSQ~\cite{esser-lsq-arxiv-2019}, where $\boldsymbol{s}$ is a learnable parameter. Therefore, the loss function of a quantized model is given as follows:
\begin{equation}\label{eqt:Loss_example}
 \mathop{\arg\min}_{\boldsymbol{w},\boldsymbol{s}} \ \  \mathbb{E}_{\boldsymbol{x}\sim \mathscr{D}_{t}} [L\left (\boldsymbol{x}; \boldsymbol{w}, \boldsymbol{s} \right)],
\end{equation}
where $L\left (\cdot;\cdot\right)$ is the predefined loss function. 

Notice that, the zero-points $\boldsymbol{z}$ are initialized by \eqref{eqt:zero_init} with the calibration set $\mathscr{D}_{c}$, and are fixed throughout the entire QAT training process. 
During QAT training, the weights involved in the forward propagation are actually the quantized weights $\hat{\boldsymbol{w}}$, rather than the floating-point weights $\boldsymbol w$. 

\subsection{Consistency Regularization}\label{sec:UCR}

\begin{figure}[t!]
    \centering
    \includegraphics[width=0.7\linewidth]{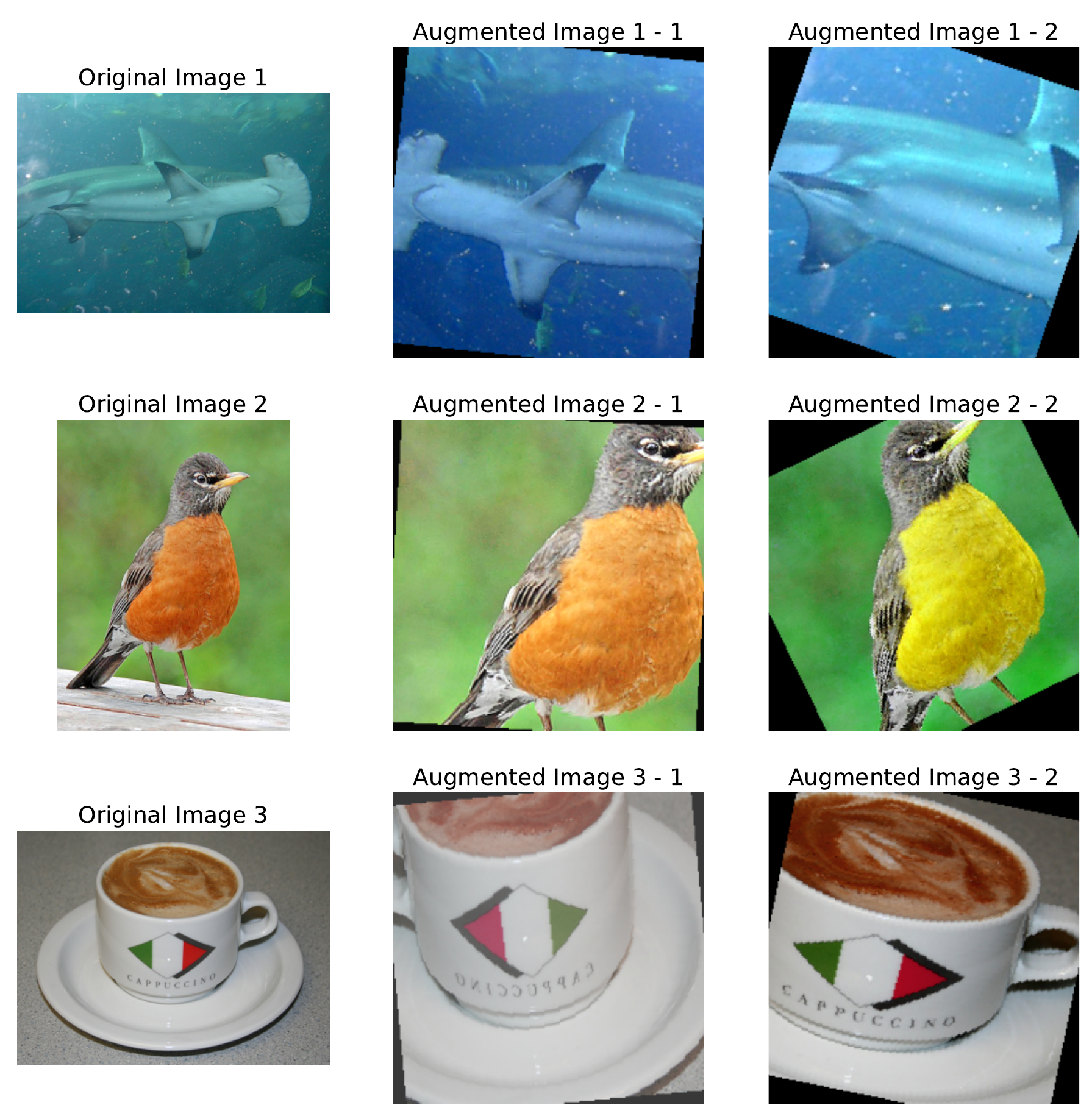}
    \caption{The augmented sample.}
    \label{fig:data_example}
\end{figure}

Given a labeled sample $\boldsymbol{x}_l$ and the corresponding label $\boldsymbol y$ (an unlabeled sample is denoted as $\boldsymbol{x}_u$, if we could easily harvest the in-distribution unlabeled data),  each sample $ \boldsymbol{x}$\footnote{We would omit the subscript if it does not cause possible confusion.} is augmented into two distinct views as follows:
\begin{eqnarray}
     \boldsymbol{x}^1_l , \boldsymbol{x}^2_l &\leftarrow DataAug\left (  \boldsymbol{x}_l ;\theta \right ),\label{eqt:dataargumented_label}
\end{eqnarray}
where $DataAug(\boldsymbol{x}_l;\theta)$ is the data augmentation function controlled by the parameter $\theta$ that determines the types of data augmentation (\textit{e.g.}, horizontal flip, random translation) and their strength (please refer Section~\ref{sec:parameter-configuration} for details). Fig.~\ref{fig:data_example} shows some images and their augmented samples. 

The framework of CR in Fig.~\ref{fig:qat-kd-cr-comparision}(c) consists of both the student model $f_S\left (\boldsymbol{x} ; \boldsymbol{w}_{S}, \boldsymbol{s}_S \right )$ and the teacher one $f_T\left (\boldsymbol{x} ; \boldsymbol{w}_{T}, \boldsymbol{s}_T \right )$. The proposed method is updated as follows:
\begin{eqnarray}\label{eqt:cr-quantization}
\begin{aligned}
f_S(\boldsymbol{x};\boldsymbol{w}_S, \boldsymbol{s}_S, \boldsymbol{z}_S) &\leftarrow QAT, \label{eqt:QAT}\\
\boldsymbol{w}_{T}^{t}&=\alpha  \boldsymbol{w}_{T}^{t-1} + \boldsymbol{w}_{S}^{t},\label{eqt:weight_ema}\\
\boldsymbol{s}_T^t&=\alpha  \boldsymbol{s}_T^{t-1} + \boldsymbol{s}_S^{t},\label{eqt:scale_ema}
\end{aligned}
\end{eqnarray}
where $QAT$ is any method (\textit{e.g.}, LSQ~\cite{esser-lsq-arxiv-2019}) to quantize weight and activation by tuning parameters $\boldsymbol{w}_S, \boldsymbol{s}_S$, $t$ is the index of mini-batch, and $\alpha$ ($\alpha > 0$) is a smoothing weight. 
The parameters $\boldsymbol{w}_{T}$ and $\boldsymbol{s}$ of the teacher model are obtained by Exponential Moving Average (EMA) in~\eqref{eqt:weight_ema}. EMA empowers the teacher model to handle the weight oscillations phenomenon~\cite{nagel-ooq-arxiv-2022}. In this paper, we set $\alpha=0.999$, as well as $\boldsymbol{s}_S$ and $\boldsymbol{s}_T$ are initialized with the same calibration dataset. 

The predictions of both the student model and the teacher one are as follows:
\begin{eqnarray}
     \boldsymbol{z}_{\boldsymbol{x}^1_l} = f_S\left (\boldsymbol{x}^1_l ; \hat{\boldsymbol{w}}_{S}, \boldsymbol{s}_S\right ),  \label{eqt:forward1}\\ 
     \boldsymbol{z}_{\boldsymbol{x}^2_l} =f_T\left (\boldsymbol{x}^2_l ;  \hat{\boldsymbol{w}}_{T}, \boldsymbol{s}_T\right ),\label{eqt:forward2}
\end{eqnarray}
where $\boldsymbol{z}_{\boldsymbol{x}^1_l}$ and $\boldsymbol{z}_{\boldsymbol{x}^2_l}$ are the outputs of the student model and the teacher one, respectively. $\hat{\boldsymbol{w}}_{S}$ and $\hat{\boldsymbol{w}}_{T}$ are the quantized weights of the student model and the teacher one, respectively.

We impose consistency loss between the outputs of the two branches for the same image $\boldsymbol{x}_l$ as follows:
\begin{equation}\label{eqt:crloss}
L_{CR} \left ( \boldsymbol{x}_l;\hat{\boldsymbol{w}}_S,\hat{\boldsymbol{w}}_{T} \right ) =KL\left ( \boldsymbol{z}_{\boldsymbol{x}^1_l}, \boldsymbol{z}_{\boldsymbol{x}^2_l}  \right ),
\end{equation}
where $KL( \cdot, \cdot )$ is Kullback-Leibler (KL) divergence for classification task. A more detailed reason would be discussed in Section~\ref{sec:reason}.  

\begin{figure}[t!]
    \centering
    \includegraphics[width=0.6\linewidth]{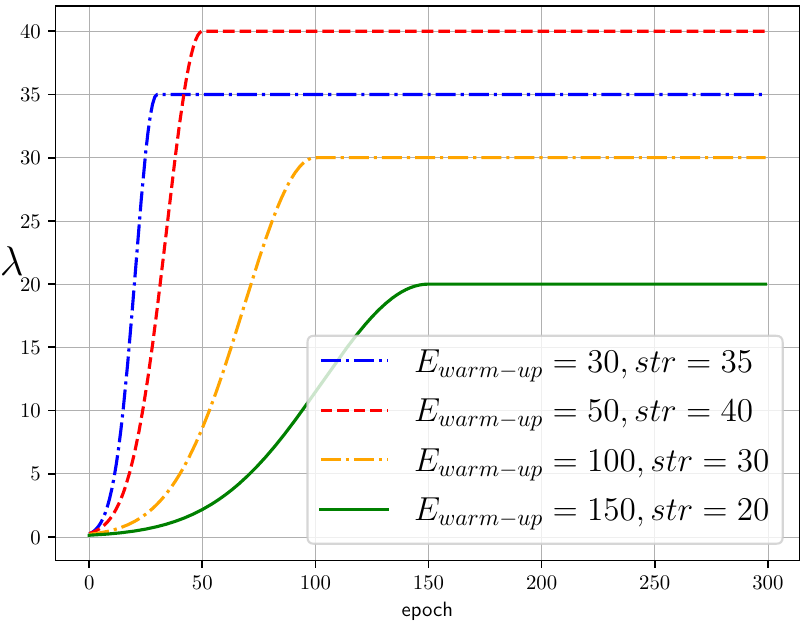}
    \caption{Different settings of the strength $\lambda$ of CR.}
    \label{fig:lambda}
\end{figure}

\subsection{Training Processing}\label{sec:training}


The loss consists of two components as follows (\textit{e.g.}, we take the classification task as an example.):
\begin{small}
\begin{equation}\label{eqt:compositedloss}
L_{\mathscr{D}_{t}}\left (\boldsymbol{x}_l;\boldsymbol{w}_S,\boldsymbol{s}_S,  \boldsymbol{w}_T, \boldsymbol{s}_T \right ) = CE\big ( \boldsymbol{z}_{\boldsymbol{x}^1_l}, \boldsymbol{y} \big ) + \lambda  L_{CR} \left ( \boldsymbol{x}_l;\hat{\boldsymbol{w}}_S,\hat{\boldsymbol{w}}_{T} \right ),
\end{equation}
\end{small}
where $\boldsymbol y$ is the ground truth, $CE\left ( \cdot ,\cdot  \right )$ denotes the cross-entropy loss for classification tasks, $\boldsymbol{y}$ denotes the ground truth, and weight $\lambda$ ($\lambda>0$) balances between the CE loss and the CR loss in~\eqref{eqt:crloss}. 

The weight $\lambda$ is critical to~\eqref{eqt:compositedloss}. We have observed that the teacher model's performance significantly depends on the student one. In this paper, $\lambda$ is progressively increased as follows:
\begin{equation}\label{eqt:lambda}
 \lambda =str\times e^{-5\times \left [ 1-\left ( \frac{\beta}{E_{\text{warm-up}}}  \right )^{2}   \right ] },
\end{equation}
where $str$ represents the maximal intensity of CR, $E_{\text{warm-up}}$ represents the predetermined of warm-up epoch, and $\beta$ is:
\begin{equation}\label{eqt:betainlambda}
 \beta= clip\left ( t,0,E_{\text{warm-up}} \right ),
\end{equation}
where $t$ represents the index of epoch.

The strength of CR is critical to the training process. Fig.~\ref{fig:lambda} illustrates that 
$\lambda$ gradually increases and stabilizes at the later stage of training. This process ensures that the model primarily optimizes the flatness of the loss landscape at the later training stage. Alg.~\ref{alg:cr} summarized the training process.

\begin{algorithm}[t!]
   \caption{QAT with CR}
   \label{alg:cr}
\begin{algorithmic}[1]
   \STATE {\bfseries Input:} 
      labeled data $\boldsymbol{x}_l$,(unlabeled data $\boldsymbol{x}_u$);
      FP model $ f\left (\boldsymbol{x} ; \boldsymbol{w}\right ) $;
      iteration $T$.
    \STATE Calibrate a FP model to initialize the student model and the teacher one;
   \FOR{$t=1$ {\bfseries to} $T$}
        \STATE Feed the augmented samples into the student model and the teacher model in~\eqref{eqt:forward1} and~\eqref{eqt:forward2};
        \STATE Optimize the loss in~\eqref{eqt:compositedloss};
        \STATE Update the student model and the teacher model in~\eqref{eqt:cr-quantization};
    \ENDFOR 
    \STATE {\bfseries Output:} 
      $f_S\left (\boldsymbol{x} ; \boldsymbol{w}_{S}, \boldsymbol{s}_S, \boldsymbol{z}_S\right )$; $f_T\left (\boldsymbol{x} ; \boldsymbol{w}_{T}, \boldsymbol{s}_T, \boldsymbol{z}_T\right )$.
\end{algorithmic}
\end{algorithm}

\subsection{Why CR Improve the Generalization of QAT?}\label{sec:reason}

\newtheorem{myobr}{Observation}
\newtheorem{mydef}{Definition}
\newtheorem{mythe}{Theorem}
\newtheorem{mypro}{Proposition}

\begin{mypro}[CR is equal to optimize flatness in~\eqref{eqt:flatness_loss_decomposition}]
\label{prop:cr-is-flatness}
If KL divergence is used in CR loss~\eqref{eqt:crloss} and CE is used in~\eqref{eqt:compositedloss}, optimizing CR loss~\eqref{eqt:crloss} is equal to minimize the flatness of a network, \textit{i.e.}, 
\begin{equation}
\arg\min_{\boldsymbol{w}} L_{\mathscr{D}_{t}}(\boldsymbol{w} + \boldsymbol{\epsilon}) -L_{\mathscr{D}_t}(\boldsymbol{w}).
\end{equation}
\end{mypro}

\begin{proof} 
Without loss of generality, when a sample $\boldsymbol{x}$ is augmented into two views $\boldsymbol{x}^1$ and $\boldsymbol{x}^2$, we assume that $\boldsymbol{x}^1= \boldsymbol{x}^2 + \boldsymbol{\epsilon}_{img}$, where $\boldsymbol{\epsilon}_{img}$ is the difference between two augmented samples. We assume that the distributions of the logits from the student model and the teacher one are $p_S(\boldsymbol x^2+\boldsymbol{\epsilon}_{img};\boldsymbol{w}_S, \boldsymbol{s}_S)$ and $p_T(\boldsymbol x^2;\boldsymbol{w}_T, \boldsymbol{s}_T)$, respectively. If KL divergence has a local minima, we have:
\begin{equation}\label{eqt:input-pertubation}
p_S(\boldsymbol x^2+\boldsymbol{\epsilon}_{img};\boldsymbol{w}_S, \boldsymbol{s}_S)\approx 
p_T(\boldsymbol x^2;\boldsymbol{w}_T, \boldsymbol{s}_T).
\end{equation}
For the same sample $\boldsymbol{x}$, two views $\boldsymbol{x}^1$ and $\boldsymbol{x}^2$ have the same label $\boldsymbol{y}$, the difference of CE losses between two views are as follows:
\begin{small}
\begin{eqnarray}\nonumber
\Delta &= &L_{\mathscr{D}_{t}}\left( \boldsymbol{x}^2+\boldsymbol{\epsilon}_{img};\boldsymbol{w}_S, \boldsymbol{s}_S \right) - L_{\mathscr{D}_{t}}\left(\boldsymbol{x}^2; \boldsymbol{w}_S, \boldsymbol{s}_S \right)\\\nonumber
&\approx&-\boldsymbol{y}\log p_S(\boldsymbol x^2+\boldsymbol{\epsilon}_{img};\boldsymbol{w}_S, \boldsymbol{s}_S)
+ \boldsymbol{y}\log p_T(\boldsymbol x^2;\boldsymbol{w}_T, \boldsymbol{s}_T)\\\nonumber
&=&0.
\end{eqnarray}
\end{small}

Based on Proposition~\ref{prop:perturbation-propagation} and~\ref{prop:perturbation_interchangeable},~\eqref{eqt:input-pertubation} could be equal to:
\begin{equation}
\arg \min_{\boldsymbol{w}} L_{\mathscr{D}_{t}}\left( \boldsymbol{x}^2;\boldsymbol{w}+\boldsymbol{\epsilon}, \boldsymbol{s}_S \right) - L_{\mathscr{D}_{t}}\left(\boldsymbol{x}^2; \boldsymbol{w}, \boldsymbol{s}_S\right),
\end{equation}
where $\boldsymbol{\epsilon}$ is determined by the multiplication of Lipschitz constant in Proposition~\ref{prop:perturbation-propagation}.

 Naturally, CR is used to minimize the flatness in~\eqref{eqt:flatness_loss_decomposition} to increase the generalization ability.  
\end{proof}

We notice that the KL divergence in~\eqref{eqt:flatness_loss_decomposition} does not involve the label information. Therefore, unlabled data could be leveraged to improve the performance of QAT. Given a in-distribution unlabeled data $\boldsymbol{x}_u$,  we could augment $\boldsymbol{x}_u$ into two distinct views as follows:
\begin{eqnarray}
     \boldsymbol{x}^1_u , \boldsymbol{x}^2_u &\leftarrow DataAug\left (  \boldsymbol{x}_u ;\theta \right ).\label{eqt:dataargumented_unlabel}
\end{eqnarray} 
The predictions of both the student model and the teacher one are as follows:
\begin{eqnarray}
     \boldsymbol{z}_{\boldsymbol{x}^1_u} = f_S\left (\boldsymbol{x}^1_u ; \hat{\boldsymbol{w}}_{S}, \boldsymbol{s}_S\right ),  \label{eqt:forward_unlabel1}\\ 
     \boldsymbol{z}_{\boldsymbol{x}^2_u} =f_T\left (\boldsymbol{x}^2_u ;  \hat{\boldsymbol{w}}_{T}, \boldsymbol{s}_T\right ),\label{eqt:forward_unlabel2}
\end{eqnarray}
where $\boldsymbol{z}_{\boldsymbol{x}^1_u}$ and $\boldsymbol{z}_{\boldsymbol{x}^2_u}$ are the outputs of student model and the teacher model, respectively. We also can impose consistency loss between the outputs for the two different views of the same unlabeled image $\boldsymbol{x}_u$ as follows:
\begin{equation}\label{eqt:crloss_unlabel}
L_{CR} \left ( \boldsymbol{x}_u;\hat{\boldsymbol{w}}_S,\hat{\boldsymbol{w}}_T \right ) =KL\left ( \boldsymbol{z}_{\boldsymbol{x}^1_u}, \boldsymbol{z}_{\boldsymbol{x}^2_u}  \right ).
\end{equation}
The influence of the unlabeled data will be discussed in Section~\ref{sec4.1}.

\begin{mypro}[generalization]
\label{prop:generalization}
For $ \rho > 0$ and any distribution $\mathscr{D}$, with probability $1-\delta$ over the choice of the training set $\mathscr{D}_{t}\sim \mathscr{D}$, a $\boldsymbol{\epsilon}$ ($\|\boldsymbol{\epsilon}\|_2 \leq \rho$) makes the following inequality hold,
\begin{small}
\begin{equation}\label{eqt:sharpness_theorem}
L_{\mathscr{D}}(\boldsymbol{w}) \leq  L_{\mathscr{D}_{t}}(\boldsymbol{w} + \boldsymbol{\epsilon}) + h(\|\boldsymbol{w}\|_2^2/\rho^2) 
\end{equation}
\end{small}
where $h:\mathcal{R}^+\rightarrow \mathcal{R}^+$ is a strictly increasing function (under some technical conditions on $ L_{\mathscr{D}}(\boldsymbol{w})$. 
\end{mypro}

The condition \( L_{\mathscr{D}}(\boldsymbol w) \leq \mathbb{E}_{\epsilon_i \sim \mathcal{N}(0,\rho)}[L_{\mathscr{D}}(\boldsymbol{w} + \epsilon)] \) means that adding Gaussian perturbation should not decrease the test error. This is expected to hold in practice for the final solution but does not necessarily hold for any \( \boldsymbol{w} \).

\begin{proof}

The proof technique we use here is inspired from~\cite{chatterji-intriguing-2019-arxiv}, PAC-Bayesian generalization bound~\cite{mcallester-pac-1999-clt} and \cite{dziugaite-computing-2017-arxiv}. We follow the conclusion in~\cite{foret-sam-iclr-2021} as follows: 
\begin{small}
\begin{equation}
\label{eqt:sharpness_proof6}
\begin{aligned}
\mathbb{E}_{\epsilon_{i} \sim \mathcal{N}(0, \sigma)}\left[L_{\mathscr{D}}(\boldsymbol{w}+ 
\boldsymbol{\epsilon})\right] \leq \mathbb{E}_{\epsilon_{i} \sim \mathcal{N}(0, \sigma)}\left[L_{\mathscr{D}_{t}}(\boldsymbol{w}+\boldsymbol{\epsilon})\right] & \\
+ \sqrt{\frac{\frac{1}{4} k \log \left(1+\frac{\|\boldsymbol{w}\|_{2}^{2}}{k \sigma^{2}}\right)+\frac{1}{4}+\log \frac{n}{\delta}+2 \log (6 n+3 k)}{n-1}}.
\end{aligned}
\end{equation}
\end{small}

In the above bound, we have \(\epsilon_i \sim \mathcal{N}(0, \sigma)\). Therefore, \(\|\boldsymbol{\epsilon}\|_2^2\) has chi-square distribution and by Lemma 1 in \cite{laurent-adaptive-2000-as}, we have that for any positive \(t\):

\begin{equation}
P\left(\|\boldsymbol{\epsilon}\|_2^2 - k\sigma^2 \geq 2\sigma^2\sqrt{kt} + 2\sigma^2 t\right) \leq \exp(-t)
\end{equation}

Therefore, with probability \(1 - \frac{1}{\sqrt{n}}\) we have that:

\begin{equation}
\begin{aligned}
\|\boldsymbol{\epsilon}\|_2^2 &\leq \sigma^2 \left(2 \ln(\sqrt{n}) + k + 2\sqrt{k \ln(\sqrt{n})}\right) \\
&\leq \sigma^2 k \left(1 + \frac{\ln(n)}{k}\right)^2 \leq\rho^2
\end{aligned}
\end{equation}

Substituting the above value for \(\sigma\) back to the inequality and using theorem's assumption, a $\boldsymbol{\epsilon}$ who satisfies $\|\boldsymbol{\epsilon}\|_2 \leq \rho$ makes the following inequality:
\begin{tiny}
\begin{equation}
\begin{aligned}
L_{\mathscr{D}}(\boldsymbol{w}) &\leq (1 - \frac{1}{\sqrt{n}}) L_{\mathscr{D}_{t}}(\boldsymbol{w} + \boldsymbol{\epsilon}) + \frac{1}{\sqrt{n}} \\
&\quad + \sqrt{
\frac{\left [ \frac{1}{4} k \log \left(1 + \frac{\|\boldsymbol{w}\|_2^2}{\rho^2}\right) \left(1 + \frac{\log(n)}{k}\right)^2 \right ] }{n-1} 
+ \frac{\left [ \log \frac{n}{\delta} + 2 \log (6n + 3k) \right ] }{n-1}} \\
&\leq L_{\mathscr{D}_{t}}(\boldsymbol{w} + \boldsymbol{\epsilon}) \\
&\quad + \underbrace{ \sqrt{
\frac{\left [ k \log \left(1 + \frac{\|\boldsymbol{w}\|_2^2}{\rho^2}\right) \left(1 + \frac{\log(n)}{k}\right)^2 \right]}{n-1} 
+ \frac{\left [4 \log \frac{n}{\delta} + 8 \log \left(6n + 3k\right) \right] }{n-1} 
}}_{h(\|\boldsymbol{w}\|_2^2)}
\end{aligned}
\end{equation}
\end{tiny}
\end{proof}

we can rewrite the inequality in~\eqref{eqt:sharpness_theorem} as follows: 
\begin{small}
\begin{equation}\label{eqt:flatness_loss_decomposition}
 L_{\mathscr{D}}(\boldsymbol{w}) \leq  L_{\mathscr{D}_t}(\boldsymbol{w}) +
 \big( L_{\mathscr{D}_{t}}(\boldsymbol{w} + \boldsymbol{\epsilon}) -L_{\mathscr{D}_t}(\boldsymbol{w})\big) +h(\|\boldsymbol{w}\|_2^2/\rho^2),  
\end{equation}
\end{small}
where $ L_{\mathscr{D}}(\boldsymbol{w})$ is \textit{any} conventional loss, \textit{e.g.}, CE loss, and $ L_{\mathscr{D}_{t}}(\boldsymbol{w} + \boldsymbol{\epsilon}) -L_{\mathscr{D}_t}(\boldsymbol{w})$ measures the flatness of the loss landscape of a neural network. In our work, when an image $\boldsymbol{x}$ is augmented into a new one $\boldsymbol{x}^\prime$, we could safely assume that $\boldsymbol{x}^\prime =\boldsymbol{x}+\boldsymbol{\epsilon}_{img}$, where $\boldsymbol{\epsilon}_{img}$ is the perturbation on the image $\boldsymbol{x}$. 

\begin{mypro}[The perturbation in the input space is interchangeable with the one in the weight space] 
\label{prop:perturbation_interchangeable}
Given an input \( \boldsymbol x \) and a neural network with the weight \( \boldsymbol w \), if the input \( \boldsymbol x \) undergoes a perturbation \( \boldsymbol{\epsilon}_{img} \), the perturbation \( \boldsymbol{\epsilon}_{img} \) is interchangeable with the perturbation \( \boldsymbol{\epsilon} \) to the weight \( \boldsymbol{w} \).
\end{mypro}

\begin{proof}
The proof technique we use here is inspired from~\cite{wei-qdrop-arxiv-2022}. We follow the conclusion in~\cite{wei-qdrop-arxiv-2022} as follows:
\begin{equation}\label{eqt:transformation}
\begin{aligned}
\mathbb{E}_{\boldsymbol{x}\sim \mathscr{D}_{t}}\left[ {L\left( \boldsymbol{x} \odot \left(1 + \boldsymbol{\epsilon}_{x} \right);\boldsymbol{w}\right) - L\left(\boldsymbol{x};\boldsymbol{w} \right)} \right] &= \\
\mathbb{E}_{\boldsymbol{x}\sim \mathscr{D}_{t}}\left[ {L\left( \boldsymbol{x};\boldsymbol{w} \odot \left( {1 + \boldsymbol{\epsilon}_{w}} \right) \right) - L\left(\boldsymbol{x}; \boldsymbol{w} \right)} \right],
\end{aligned}
\end{equation}
where $\odot$ is the point-wise multiplication, and $\boldsymbol{\epsilon}_{x}$ and $\boldsymbol{\epsilon}_{w}$ could be \textit{any} perturbation from the input and the weight, respectively. 
We can rewrite~\eqref{eqt:transformation} as follows:
\begin{equation}
\begin{aligned}
\mathbb{E}_{\boldsymbol{x}\sim \mathscr{D}_{t}}\left[ {L\big( \boldsymbol{x} +\underbrace{ \boldsymbol{x}\odot \boldsymbol{\epsilon}_x }_{\boldsymbol{\epsilon}_{img} } ;\boldsymbol{w}\big) - L\left(\boldsymbol{x};\boldsymbol{w} \right)} \right] &= \\
\mathbb{E}_{\boldsymbol{x}\sim \mathscr{D}_{t}}\left[ {L\big( \boldsymbol{x};\boldsymbol{w} + \underbrace{\boldsymbol{w} \odot \boldsymbol{\epsilon}_{w}}_{\boldsymbol{\epsilon}} \big) - L\left(\boldsymbol{x}; \boldsymbol{w} \right)} \right],
\end{aligned}
\end{equation}
Without loss of generality, we could find a $\boldsymbol{\epsilon}_x$ to make $\boldsymbol{\epsilon}_{img}= \boldsymbol{x}\odot \boldsymbol{\epsilon}_x$ hold; while, we could also find a $\boldsymbol{\epsilon}_{w}$ to make $\hat{\boldsymbol{w}} \odot \boldsymbol{\epsilon}_{w} = \boldsymbol{\epsilon}$ 
\end{proof}

\begin{mypro}[Expansion of perturbation from the input to all layers of a network]
\label{prop:perturbation-propagation}
Without loss of generality, we assume a feed-forward network $f(\boldsymbol{x})$ composed of a serial of layers $\phi_{i}$:
\begin{equation}
\label{eqt:feed-forward}
f(\boldsymbol{x})=\left(\phi_{L} \ldots \circ \phi_{l} \circ \ldots \circ \phi_{0}\right)(\boldsymbol{x}).
\end{equation}
Given the perturbation $\boldsymbol{\epsilon}_{img}$ from the input $\boldsymbol{x}$, and the Lipschitz constant of the layer $ Lip(\phi_{l})$,  if the multiplication $\prod_{l=0}^L Lip(\phi_{l})$ is not approximately to zero, the perturbation $\boldsymbol{\epsilon}_{img}$ would propagate across the all $L$ layers of a network.
\end{mypro}
 $\phi_{i}$ can be a linear layer, conventional layer, pooling, activation functions, \textit{etc}.

 
\begin{proof}
We assume that the Lipschitz constant of the $i$-th layer $\phi_{i}$ in a network is $Lip(f)$. We consider a network $f$ with the input $\boldsymbol{x}$ and the perturbed input as $\boldsymbol{x}'=\boldsymbol{x}+\boldsymbol{\epsilon}_{img}$. 
According to the Lipschitz condition, we have:
\begin{equation}\label{eqt:fist-layer-lipschitz}
|\phi_0(\boldsymbol{x}') - \phi_0(\boldsymbol{x})| \leq Lip(\phi_0) \cdot |\boldsymbol{\epsilon}_{img}|.
\end{equation}

\eqref{eqt:fist-layer-lipschitz} indicates that if the Lipschitz constant of the first layer is non-zero, the perturbation~\(\boldsymbol{\epsilon}_{img}\) in the input layer will be propagated to the output of the first layer with the scale \(Lip(\phi_0)\). Specifically, if \(Lip(\phi_0) > 1\), the input perturbation of the first layer will be amplified; if \(Lip(\phi_0) < 1\), the input error will be attenuated.

For any $l$-th layer of the network, we have:
\begin{equation}
\begin{aligned}
\label{eqt:pertubation_amplify}
&\quad |\phi_L(\cdots \phi_1(\phi_0(\boldsymbol{x}')) \cdots)| - |\phi_L(\cdots \phi_1(\phi_0(\boldsymbol{x})) \cdots)| \\
&\leq Lip(\phi_L) \cdot |\phi_{L-1}(\cdots \phi_0(\boldsymbol{x}')) - \phi_{L-1}(\cdots \phi_0(\boldsymbol{x}))| \\
&\leq Lip(\phi_L) \cdot Lip(\phi_{L-1}) \cdots Lip(\phi_0) \cdot |\boldsymbol{\epsilon}_{img}| \\
& \leq m \cdot|\boldsymbol{\epsilon}_{img}|
\end{aligned}
\end{equation}
where \(m = Lip(\phi_0) \cdots Lip(\phi_{l}) \cdots Lip(\phi_{L})\). 
\end{proof}
In practice, we have empirically found that the Lipschitz constants of neural network layers are often greater than 1~\cite{lin-defensive-iclr-2019}, \cite{cisse-parseval-icml-2017}. Therefore, during the quantization process of neural networks, we do not check or regularize that whether the Lipschitz constant of each layer is larger than 1 or not.

\section{Experiments}


\textbf{Experimental Protocols.} 
Our code is based on PyTorch~\cite{paszke-pytorch-nips-2019} and relies on the MQBench~\cite{li-mqbench-arxiv-2021} package. By default, we set the $\beta$ increase linearly from 0 to 4 as the number of epochs increases unless explicitly mentioned otherwise. We use the asymmetric quantization.

In this paper, two datasets, the ImageNet, CIFAR-10 and CIFAR-100, are used. We randomly selected 1,028 images for ImageNet and 100 for CIFAR-10 as the calibration set. We also keep the first and last layers with 8-bit quantization, the same as QDrop~\cite{wei-qdrop-arxiv-2022}. Additionally, we employed per-channel quantization for weight quantization. We used WXAX to represent X-bit weight and activation quantization. 


Two experimental settings are evaluated as follows:
\begin{itemize}
\item[] (A) We leverage all labeled data and utilize the entire data distribution to in CR loss.
\item[] (B) We use 20\% labeled data to simulate a scenario with the sufficient unlabeled data.
\end{itemize}

\textbf{Training Details.} We used SGD as the optimizer, with a batch size of 256 and a base learning rate of 0.01. The default learning rate (LR) scheduler followed the cosine annealing method. The weight decay was 0.0005, and the SGD momentum was 0.9. We trained for 200 epochs on CIFAR-10 and ImageNet unless otherwise specified. $E_{warm-up}$ was set as 50, $srt$ was set as 40.

\begin{table}[t!]
\begin{center}
\centering
\caption{Ablation studies of the choice of CR (Accuracy $\%$) on CIFAR-10.}
\label{table2}
\begin{tabular}{cccc}
\toprule
\begin{tabular}[c]{@{}l@{}}
$J\left ( \boldsymbol{z}_{\boldsymbol{x}^1_l}, \boldsymbol{z}_{\boldsymbol{x}^2_l} \right )$
\end{tabular} &
\begin{tabular}[c]{@{}l@{}}$J\left ( \boldsymbol{z}_{\boldsymbol{x}^1_u}, \boldsymbol{z}_{\boldsymbol{x}^2_u}  \right )$
\end{tabular} &
\begin{tabular}[c]{@{}l@{}}ResNet-18\\
\end{tabular} &
\begin{tabular}[c]{@{}l@{}}MobileNetV2\\
\end{tabular}

\\ \midrule
                  &                               &85.89                                 &75.27                             \\\midrule
 \checkmark      &                               &87.19                                 &76.77                             \\
                 & \checkmark                    &88.10                                 &78.27                             \\
 \checkmark      & \checkmark                    &\textbf{88.56}                                 &\textbf{78.47}                             \\
\bottomrule
\end{tabular}
\end{center}
\end{table}

\subsection{Ablation Study of Labeled and Unlabeled Data}
\label{sec4.1}

\textbf{Effectiveness of Unlabeled In-Distribution Data:} We used 20,000 samples as labeled data and all the remaining samples as unlabeled data to verify the effectiveness of CR on CIFAR-10 in Tab.~\ref{table2}. The pre-trained FP models, ResNet-18~\cite{he-resnet-cvpr-2016} and MobileNetV2~\cite{sandler-mobilenetv2-cvpr-2018}, were used as the baselines. Table~\ref{table2} shown that CR have significantly improved the accuracy of ResNet and MobileNetV2. Both the unlabeled and labeled data would improve the accuracies of the W4A4 model by CR; besides, the more data were used, the more gain the CR supplied.

\begin{table}[h!]
\begin{center}
\centering
\caption{Comparison among different QAT strategies regarding accuracy on CIFAR-10 when partial data are used (setting (B)).}
\label{tbl:comp-CIFAR-10-partialdata}
\resizebox{1\linewidth}{!}{
\begin{tabular}{clrcccccc}
\toprule
\multicolumn{1}{c}{\begin{tabular}[c]{@{}c@{}}\textbf{Labeled} \\ \textbf{data}\end{tabular}} & \multicolumn{1}{l}{\textbf{Methods}} & \multicolumn{1}{c}{\textbf{W/A}} & \multicolumn{1}{c}{\textbf{Res18}} & \multicolumn{1}{c}{\textbf{Res50}} & \multicolumn{1}{c}{\textbf{Reg600M}} & \multicolumn{1}{c}{\textbf{Reg3.2G}} & \multicolumn{1}{c}{\textbf{MBV1}} & \multicolumn{1}{c}{\textbf{MBV2}} \\ \midrule
10000                       & Full Prec.           & 32/32 &77.67          &75.50           &72.39            &78.61           &76.59           &75.21      \\ \hline\hline
\multirow{10}{*}{10000}     & PACT (2018)          & 4/4   &78.38          &75.74           &71.20            &76.87           &64.60           &69.34      \\
                            & LSQ (2020)           & 4/4   &79.34          &77.65           &71.62            &79.50           &74.58           &74.89      \\
                            & LSQ+ (2020)          & 4/4   &79.61          &78.05           &71.85            &78.98           &74.30           &74.68  \\
                            & KD(2018)   & 4/4   &81.13          &80.00           &75.22            &82.74           &78.22  &\textbf{76.34}  \\
                            & CR (Ours)           & 4/4    &\textbf{83.97} &\textbf{82.56}  &\textbf{75.75}   &\textbf{84.32}  &\textbf{78.72}  &75.32     \\ \cmidrule{2-9}
                            & PACT (2018)          & 2/4   &78.99          &75.60           &68.60            &76.23           &53.34           &59.67      \\
                            & LSQ  (2020)          & 2/4   &79.06          &77.56           &70.30            &77.83           &69.00           &\textbf{73.42}    \\
                            & LSQ+ (2020)          & 2/4   &78.40          &77.39           &69.53            &78.14           &68.30           &70.71      \\
                            & KD(2018)   & 2/4   &81.23          &79.79           &\textbf{74.16}   &78.14           &72.35           &71.23  \\
                            & CR (Ours)           & 2/4    &\textbf{83.20} &\textbf{82.36}  &72.73            &\textbf{83.22}  &\textbf{73.90}  &70.04      \\\bottomrule
\end{tabular}
}
\end{center}
\end{table}

Tab.~\ref{table2} also indicates that if we could find enough unlabeled in-distribution samples, CR would significantly improve the accuracy. To valid this observation, following the setting (B), Tab.~\ref{tbl:comp-CIFAR-10-partialdata} illustrated that CR significantly outperforms the STOA methods, \textit{e.g.}, LSQ and LSQ+. For instance, CR (83.2$\%$) significantly surpasses the FP (77.67$\%$) when the W2A4 setting was used. Empirical results verified that if we could harvest enough number of unlabeled in-distribution data, CR consistently outperformed the STOAs with the different neural architectures.


\begin{table}[h!]
\label{tab:unlabeled_data}
\centering
\caption{Ablation studies on the number of unlabeled data.}
\begin{tabular}{ccc}
\toprule
The number of unlabeled data & Acc (\%) \\
\midrule
500    & 81.90 \\
1000   & 82.40 \\
5000   & 82.47 \\
10000  & 82.89 \\
20000  & 83.20 \\
30000  & 83.04 \\
40000  & \textbf{83.20} \\
\bottomrule
\end{tabular}
\end{table}

\textbf{Effectiveness of the Number of Unlabeled In-Distribution Data:} We sampled 10,000 samples from the CIFAR-10 dataset as labeled data to test the ResNet-18. We gradually increased the number of unlabeled data from 500 to 40,000. Tab.~\ref{tab:unlabeled_data} verified that unlabeled data consistently have a positive influence on performance, when the number of unlabeled data was gradually increased. However, as the number of unlabeled samples reached a certain threshold, the performance gain increased slowly.

\begin{table}[h!]
\setlength{\belowcaptionskip}{0.3cm}
\begin{center}
\centering
\caption{Ablation studies of different data augmentations on CIFAR-10.}
\label{table3}
\begin{tabular}{lc}
\toprule
\multicolumn{1}{c}{Data Aug.}  & \begin{tabular}[c]{@{}c@{}} Val Acc(\%)\end{tabular}       \\ \midrule
Full Prec. & 88.73\\ \midrule
\begin{tabular}[c]{@{}l@{}}RHF \end{tabular}                                        & 89.19                                                         \\
\begin{tabular}[c]{@{}l@{}}RHF + RT\end{tabular}                    & 89.45                                                            \\
\begin{tabular}[c]{@{}l@{}}RHF + RT + RR\end{tabular}  & 88.22                                                            \\
\begin{tabular}[c]{@{}l@{}}RHF + RT + RG\end{tabular} & 89.76                                                            \\
\begin{tabular}[c]{@{}l@{}}RHF + RT + CJ\end{tabular}     & \textbf{89.90}                                                            \\ \bottomrule
\end{tabular}
\end{center}
Random Horizontal Flip (RHF), Random Translation (RT), Random Rotation (RR), Random Grayscale (RG), and Color Jitter (CJ).
\end{table}

\subsection{Ablation Study of Data Augmentation}
\label{sec:parameter-configuration}

Data augmentation is crucial to generate the reasonable perturbation $\boldsymbol{\epsilon}_{img}$ for CR. We tested several commonly used data augmentations. We tested ResNet-18~\cite{he-resnet-cvpr-2016} and used all samples from the CIFAR-10 dataset as labeled data. Tab. \ref{table3} shows that a positive correlation between the data augmentation and the accuracy of our method on CIFAR-10. The results indicated that mixing multiple data augmentations would improve the accuracy. In the following experiments, the RHF, RT, and CJ were used for the following experiments.

\subsection{Ablation Study of CR Strength}
\label{sec:CR_strength}

\begin{table}[h]

\centering
\caption{Ablation studies of Consistency regularization strength experiment}
\label{tbl:cr_strength}
\begin{tabular}{lc}
\toprule
\textbf{CR Strength Setting} & \textbf{Val Acc(\%)} \\ 
\midrule
$E_{warm-up}$ = 0, $str$ = 40 & 89.00 \\
$E_{warm-up}$ = 50, $str$ = 40 & \textbf{89.97} \\ \hline
$E_{warm-up}$ = 30, $str$ = 35 & 89.84 \\ 
$E_{warm-up}$ = 50, $str$ = 35 & 89.83 \\ 
$E_{warm-up}$ = 100, $str$ = 35 & 89.74 \\ 
\bottomrule
\end{tabular}
\end{table}

Tab.~\ref{tbl:cr_strength} discovered two observations: 1) gradually increasing the strength $str$ is critical to have a good performance, \textit{e.g.}, without the warm-up stage, CR had a worse performance, \textit{i.e.}, 89.00 vs. 89.97; 2) the results were robust to the change of the warm-up parameter, \textit{e.g.}, when the different warm-up parameters were used, the variance was 0.12\%. At the early stage of training, the $E_{warm-up}$ significantly affects the parameters of a model; therefore, the strength $E_{warm-up}$ in CR should have a gradually increased value. When at the later training stage, the value of $\lambda$ should gradually increase and stabilize to optimize the flatness of loss landscape. Moreover, Tab.~\ref{tbl:cr_strength} demonstrated that dynamic adjustment of $E_{warm-up}$ and $str$ is critical to obtain a robust yet good performance.

\subsection{Comparisons with the SOTA methods}
\label{sec4.3}
We selected ResNet-18 and -50~\cite{he-resnet-cvpr-2016} with normal convolutions, MobileNetV1~\cite{howard-mobilentv1-arxiv-2017} and V2~\cite{sandler-mobilenetv2-cvpr-2018} with depth-wise separable convolutions, and RegNet~\cite{radosavovic-regnet-cvpr-2020} with group convolutions as the diverse network architectures. We compared our approach with the SOTA methods, \textit{i.e.}, LSQ~\cite{esser-lsq-arxiv-2019}, LSQ+~\cite{bhalgat-lsq+-cvpr-2020}, PACT~\cite{choi-pact-arxiv-2018}, and KD~\cite{hinton-kd-arxiv-2015}.

\begin{table}[t!]
\begin{center}
\centering
\caption{Comparison among different QAT strategies in terms of accuracy on CIFAR-10 when all the labeled data are used (setting (A)).}
\label{tbl:comp-CIFAR-10-alldata}
\resizebox{1\linewidth}{!}{
\begin{tabular}{clrcccccc}
\toprule
\multicolumn{1}{c}{\begin{tabular}[c]{@{}c@{}}\textbf{Labeled} \\ \textbf{data}\end{tabular}} & \multicolumn{1}{l}{\textbf{Methods}} & \multicolumn{1}{c}{\textbf{W/A}} & \multicolumn{1}{c}{\textbf{Res18}} & \multicolumn{1}{c}{\textbf{Res50}} & \multicolumn{1}{c}{\textbf{Reg600M}} & \multicolumn{1}{c}{\textbf{Reg3.2G}} & \multicolumn{1}{c}{\textbf{MBV1}} & \multicolumn{1}{c}{\textbf{MBV2}} \\ \midrule
50000                       & Full Prec.     & 32/32 & 88.72         &89.95           &82.07            &87.96           &85.52           &85.81      \\ \hline\hline
\multirow{10}{*}{50000}     & PACT (2018)          & 4/4   &88.15          &85.27           &81.09            &85.29           &80.77           &79.88      \\
                            & LSQ (2020)           & 4/4   &86.69          &90.01           &82.12            &88.42           &82.39           &84.45      \\
                            & LSQ+ (2020)          & 4/4   &88.40          &90.30           &81.85            &87.32           &84.32           &84.30  \\
                            & KD(2018)   & 4/4   &88.86          &90.34           &81.41            &87.97           &84.77           &83.79  \\
                            & CR (Ours)            & 4/4   &\textbf{90.48} &\textbf{91.30}  &\textbf{84.18}   &\textbf{90.02}  &\textbf{86.23}  &\textbf{85.65}     \\ \cmidrule{2-9}
                            & PACT (2018)          & 2/4   &87.55          &85.24           &74.49            &82.68           &69.04           &67.18      \\
                            & LSQ  (2020)          & 2/4   &88.36          &90.01           &81.02            &86.96           &78.15           &78.15      \\
                            & LSQ+ (2020)          & 2/4   &87.76          &89.62           &80.22            &86.34           &81.26           &77.00      \\
                            & KD(2018)   & 2/4   &88.83          &\textbf{90.18}  &78.53            &86.48           &78.84           &75.56  \\
                            & CR (Ours)            & 2/4   &\textbf{89.90} &90.13  &\textbf{81.53}   &\textbf{88.27}  &\textbf{82.86}  &\textbf{80.22}      \\\bottomrule

\end{tabular}
}
\end{center}
\end{table}

\textbf{CIFAR-10.} Tab.~\ref{tbl:comp-CIFAR-10-alldata} illustrates that when the whole training set of CIFAR-10 was used as the labeled samples, CR significantly surpassed these SOTAs. For W4A4 quantization, CR achieved about 1$\sim$3$\%$ accuracy improvements over LSQ. Furthermore, for W2A4 quantization, CR consistently achieved a 1$\sim$3$\%$ accuracy improvement over LSQ in Tab.~\ref{tbl:comp-CIFAR-10-alldata}. 

Even when the number of labeled sampled was reduced in Tab.~\ref{tbl:comp-CIFAR-10-partialdata}, CR still exhibited an excellent performance, \textit{e.g.}. For instance, the W4A4 setting in Tab.~\ref{tbl:comp-CIFAR-10-partialdata} shows that CR achieved a 4$\%$ accuracy improvement over ResNet and a 2$\%$ accuracy improvement over MobileNet, respectively; besides, when W2A4 setting was adopted in Tab.~\ref{tbl:comp-CIFAR-10-partialdata}, CR consistently achieved a 2$\%$ accuracy improvement over ResNet and a 4$\%$ accuracy improvement over MobileNet, respectively.

\begin{table}[t!]
\begin{center}
\centering
\caption{Comparison among different QAT strategies in terms of accuracy on CIFAR-100 when all the labeled data are used (setting (A)).}
\label{tbl:comp-CIFAR-100-alldata}
\resizebox{1\linewidth}{!}{
\begin{tabular}{clrccccc}
\toprule
\multicolumn{1}{c}{\begin{tabular}[c]{@{}c@{}}\textbf{Labeled} \\ \textbf{data}\end{tabular}} & \multicolumn{1}{l}{\textbf{Methods}} & \multicolumn{1}{c}{\textbf{W/A}} & \multicolumn{1}{c}{\textbf{Res18}} & \multicolumn{1}{c}{\textbf{Res50}} & \multicolumn{1}{c}{\textbf{Reg600M}}  & \multicolumn{1}{c}{\textbf{MBV1}} & \multicolumn{1}{c}{\textbf{MBV2}} \\ \midrule
50000                       & Full Prec.           & 32/32 & 75.4         &78.94           &76.57            &70.22           &71.30        \\ \hline\hline
\multirow{10}{*}{50000}     & PACT (2018)          & 4/4   &74.17         &74.78           &72.74            &64.65           &64.06       \\
                            & LSQ (2020)           & 4/4   &75.30         &78.20           &76.31            &68.63           &69.01        \\
                            & LSQ+ (2020)          & 4/4   &74.50         &77.39           &75.08            &67.89           &68.25        \\
                            & KD(2018)             & 4/4   &74.70         &78.80           &75.20            &\textbf{70.96}  &\textbf{71.66}        \\
                            & CR (Ours)            & 4/4   &\textbf{75.34} &\textbf{78.97} &\textbf{77.33}   &69.40           &70.50        \\ \cmidrule{2-8}
                        
                            & PACT (2018)          & 2/4   &73.77          &74.72           &75.41            &49.98           &57.90       \\
                            & LSQ  (2020)          & 2/4   &74.93          &77.82           &75.32            &65.13           &66.15        \\
                            & LSQ+ (2020)          & 2/4   &73.90          &76.61           &74.15            &65.28           &66.24         \\
                            & KD(2018)             & 2/4   &74.35          &77.34           &74.80            &66.90           &63.77         \\
                            & CR (Ours)            & 2/4   &\textbf{74.89} &\textbf{78.36}  &\textbf{76.43}   &\textbf{67.35}  &\textbf{68.32}      \\\bottomrule

\end{tabular}
}
\end{center}
\end{table}

\textbf{CIFAR-100.} Tab.~\ref{tbl:comp-CIFAR-100-alldata} and Tab.~\ref{tbl:comp-CIFAR-100-partialdata} show that the different models were quantized into W4A4 and W2A4, respectively. Tab.~\ref{tbl:comp-CIFAR-100-alldata} illustrated that our method consistently outperformed the counterparts, \textit{i.e.}, PACT, LSQ, and LSQ+. For instance, CR achieved approximately 0.2$\%$ to 1$\%$ accuracy improvement over LSQ in W4A4 quantization. However, our method achieved inferior results than that of KD on the light-weight network architectures, \textit{i.e.}, MobileNetV1 and MobileNetV2. We provide a detailed explanation of this phenomenon in Section~\ref{sec:generalization}.

Tab.~\ref{tbl:comp-CIFAR-100-partialdata} adopted 60$\%$ labeled samples (\textit{i.e.}, 30000 samples) to verify the effectiveness of unlabled data. CR significantly surpassed the STOA methods. For instance, CR significantly improved the performances over the STOA methods by leveraging unlabeled data to optimize the flatness, \textit{e.g.}, CR achieved about 1.5$\%$ to 3$\%$ accuracy improvements over other methods for W4A4 quantization; while, for W2A4 quantization, CR improved MobileNet’s performance from 66.24$\%$ to 68.32$\%$.

\begin{table}[t!]
\begin{center}
\centering
\caption{Comparison among different QAT strategies in terms of accuracy on CIFAR-100 when partial data are used (setting (B)).}
\label{tbl:comp-CIFAR-100-partialdata}
\resizebox{1\linewidth}{!}{
\begin{tabular}{clrccccc}
\toprule
\multicolumn{1}{c}{\begin{tabular}[c]{@{}c@{}}\textbf{Labeled} \\ \textbf{data}\end{tabular}} & \multicolumn{1}{l}{\textbf{Methods}} & \multicolumn{1}{c}{\textbf{W/A}} & \multicolumn{1}{c}{\textbf{Res18}} & \multicolumn{1}{c}{\textbf{Res50}} & \multicolumn{1}{c}{\textbf{Reg600M}}  & \multicolumn{1}{c}{\textbf{MBV1}} & \multicolumn{1}{c}{\textbf{MBV2}} \\ \midrule
30000                       & Full Prec.           & 32/32 & 71.7         &70.75           &69.68            &61.68           &65.89        \\ \hline\hline
\multirow{10}{*}{30000}     & PACT (2018)          & 4/4   &71.67         &69.95           &69.37            &62.31           &59.98       \\
                            & LSQ (2020)           & 4/4   &73.26         &73.56           &70.92            &63.08           &65.86        \\
                            & LSQ+ (2020)          & 4/4   &72.90         &72.84           &70.61            &61.34           &66.17        \\
                            & KD(2018)             & 4/4   &72.19         &72.33           &70.05            &62.97           &66.31       \\
                            & CR (Ours)            & 4/4   &\textbf{74.84} &\textbf{75.15} &\textbf{73.88}   &\textbf{64.42}  &\textbf{68.16}         \\ \cmidrule{2-8}
                        
                            & PACT (2018)          & 2/4   &71.01          &69.48           &68.57            &44.70           &55.16        \\
                            & LSQ  (2020)          & 2/4   &72.78          &73.04           &70.21            &60.88           &64.74        \\
                            & LSQ+ (2020)          & 2/4   &72.53          &72.73           &69.37            &61.75           &64.59         \\
                            & KD(2018)             & 2/4   &72.45          &72.09           &70.59            &61.06           &63.07         \\
                            & CR (Ours)            & 2/4   &\textbf{74.02} &\textbf{74.66}  &\textbf{72.59}   &\textbf{63.30}  &\textbf{65.95}      \\\bottomrule

\end{tabular}
}
\end{center}
\end{table}

\begin{table}[h!]
\begin{center}
\centering
\caption{Comparison among different QAT strategies with W4A4 regarding accuracy on ImageNet.}
\label{tbl:comp-imagenet}
\begin{tabular}{clrcccccc}
\toprule
Labeled data                    & Method         & Res18 & Res50  \\ \midrule
\multirow{5}{*}{Entire dataset} & Full Prec.     & 71.00 & 77.00   \\\hline\hline
                                & PACT (2018)    & 69.20 & 76.50   \\
                                & DSQ (2019)     & 69.56 &   -     \\
                                & LSQ (2020)     & \textbf{71.10} & 76.70  \\
                                & CR (Ours)      & 70.85     &\textbf{76.93}  \\  \bottomrule
\end{tabular}
\end{center}
\end{table}

\textbf{ImageNet.} We also investigated the effectiveness of CR on the ImageNet. We follow the setting (A) in the experimental configuration. Tab.~\ref{tbl:comp-imagenet} compared our approach with the SOTA methods. For a fair comparison, we compared the results of FP, PACT~\cite{choi-pact-arxiv-2018}, DSQ~\cite{gong-dsq-cvpr-2019}, and LSQ~\cite{esser-lsq-arxiv-2019}, respectively. Our method outperformed the results reported by LSQ, which learned the step size. For instance, CR improved Res50's performance from 76.7$\%$ to 76.93$\%$. CR pushed the W4A4 towards the FP result, \textit{i.e.}, 77.0$\%$.

\subsection{Characteristics of CR}
\label{sec:generalization}

\begin{figure}[t!]
    \centering
    \includegraphics[width=0.6\linewidth]{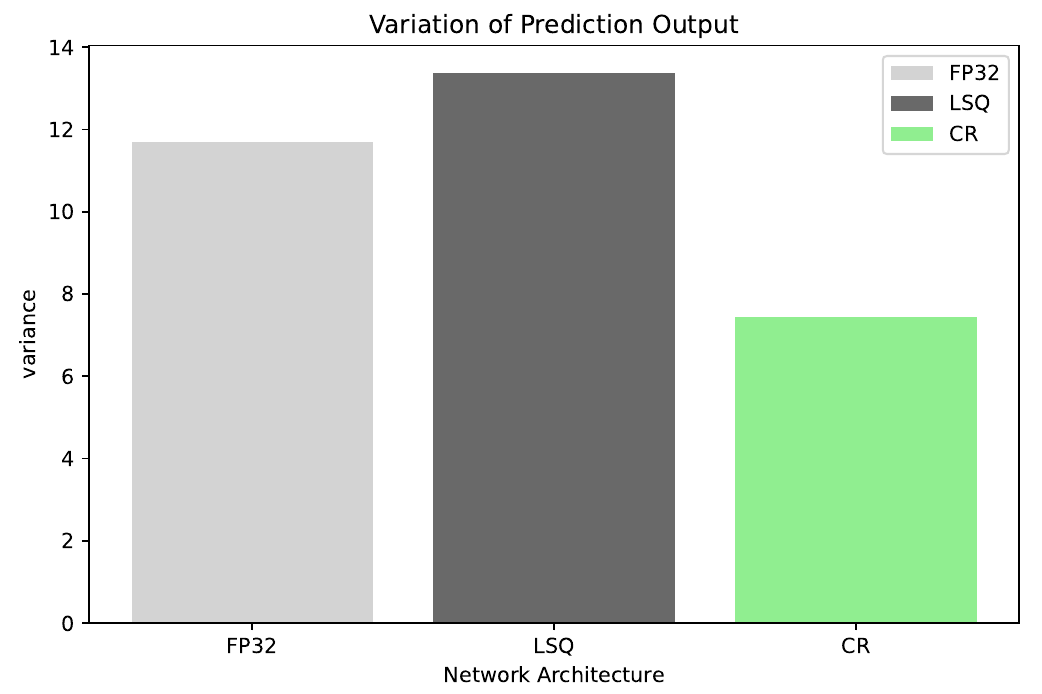}
    \caption{The output variance of the CR, LSQ, and FP32 models of ResNet18 network on CIFAR-10 dataset.}
    \label{fig:variance}
\end{figure}

\textbf{Robustness of CR:} Robustness is defined as the property that if a testing sample is ``similar'' to a training one, then the testing error is close to the training error~\cite{xu-robustness-ml-2012}. We applied Gaussian-distribution perturbations to the original inputs to investigate the robustness as follows:
\begin{equation}
\label{eqt:perturbation}
\boldsymbol{z}=f\left ( \boldsymbol{x}+\boldsymbol{\epsilon}_{img};\boldsymbol{w} \right ) ,
\end{equation}
where $\boldsymbol{x}$ represents the model input, perturbation $\boldsymbol{\epsilon}_{img}$ follows the Gaussian-distribution, $\boldsymbol{w}$ represents parameters of a model, $\boldsymbol{z}$ is the output of the model, and $f\left ( \cdot ;\cdot  \right )$ represents the FP model or the quantized one. The variance of the output $\boldsymbol{z}$ is illustrated in Fig.~\ref{fig:variance}.

\begin{figure*}[t!]
    \centering
    \includegraphics[width=1.0\linewidth]{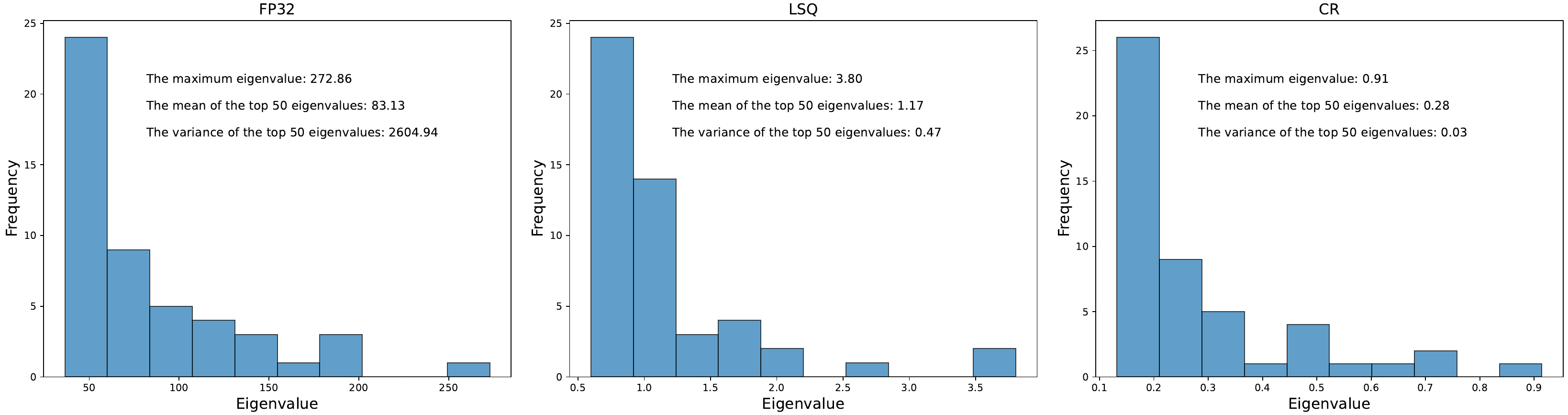}
    \caption{Frequency plot of the top 50 eigenvalues for FP32, LSQ, and CR. The plot illustrate the maximum, mean, and variance of the top 50 eigenvalues.}
    \label{fig:eigenvalues-comparision}
\end{figure*}

\begin{figure}[h!]
    \centering
    \includegraphics[width=1.0\linewidth]{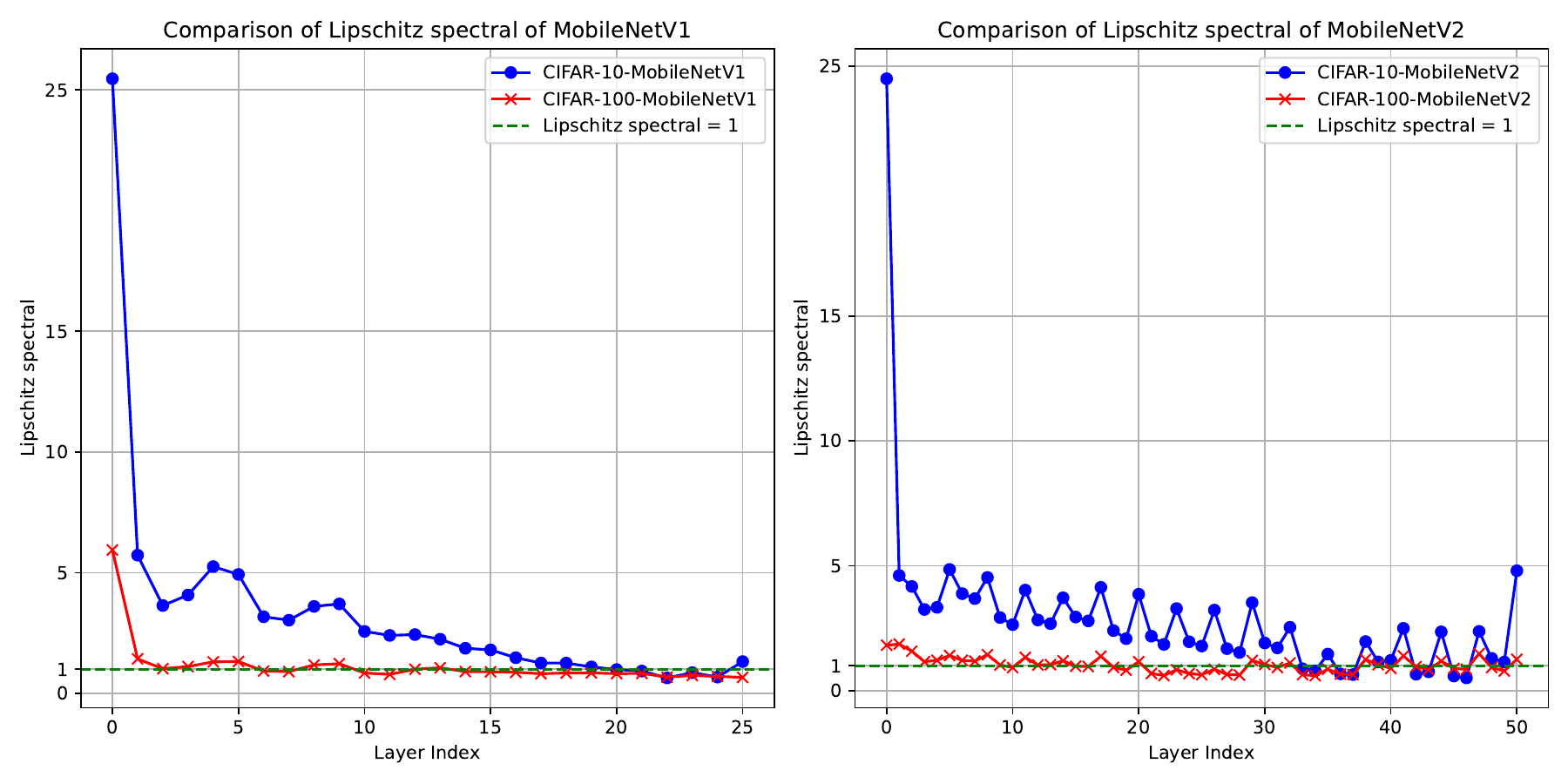}
    \caption{The Lipschitz spectral comparison between MobileNetV1 and MobileNetV2 models on the CIFAR-10 and CIFAR-100 datasets.}
    \label{fig:Lipschitz-comparison}
\end{figure}

Fig.~\ref{fig:variance} showed that the output variance of the LSQ model was larger than that of the FP model. The result indicated that the quantization process acted as a variance amplifier, magnifying the variance of the errors introduced by quantization. In contrast, our method successfully suppressed the noise from the input. The explanation is that the loss of CR would cause the perturbation $\boldsymbol{\epsilon}_{img}$ to be gradually absorbed across different layers of a network.  

Proposition~\ref{prop:perturbation-propagation} explains that if the Lipschitz constant of a layer in a network is greater than 1, the perturbation $\boldsymbol{\epsilon}_{img}$ applied to an image will be propagated as the perturbations of the weights $\boldsymbol{\epsilon}$. Therefore, we illustrated the Lipschitz constant of each layer in the FP models MobileNetV1 and MobileNetV2 trained on the CIFAR-10 and CIFAR-100 datasets, respectively. We used the method in \cite{virmaux-lipschitz-cmp-2018-nips} that considered Lipschitz spectra as a replacement for the calculation of Lipschitz constant. Fig.~\ref{fig:Lipschitz-comparison} shown that the Lipschitz spectral of the network layers in the MobileNet  was higher than that of the FP model on CIFAR-100. For instance, in the first layer of MobileNetV1, the Lipschitz spectrum for CIFAR100 is 25.46, while for CIFAR10 it is 5.95. In the first layer of MobileNetV2, the Lipschitz spectrum for CIFAR100 is 24.49, while for CIFAR10 it is 1.82. Due to the smaller Lipschitz spectral of the MobileNet on CIFAR-100, the CR method made the MobileNet on CIFAR-100 perform worse than that of the models on CIFAR-10, \textit{e.g.}, 69.4 vs. 70.22 for MobileNetV1, and 70.55 vs. 71.30 for MovileNetV2 for W4A4 in Tab.~\ref{tbl:comp-CIFAR-100-alldata}.

\begin{figure}[t!]
    \centering
    \includegraphics[width=1.0\linewidth]{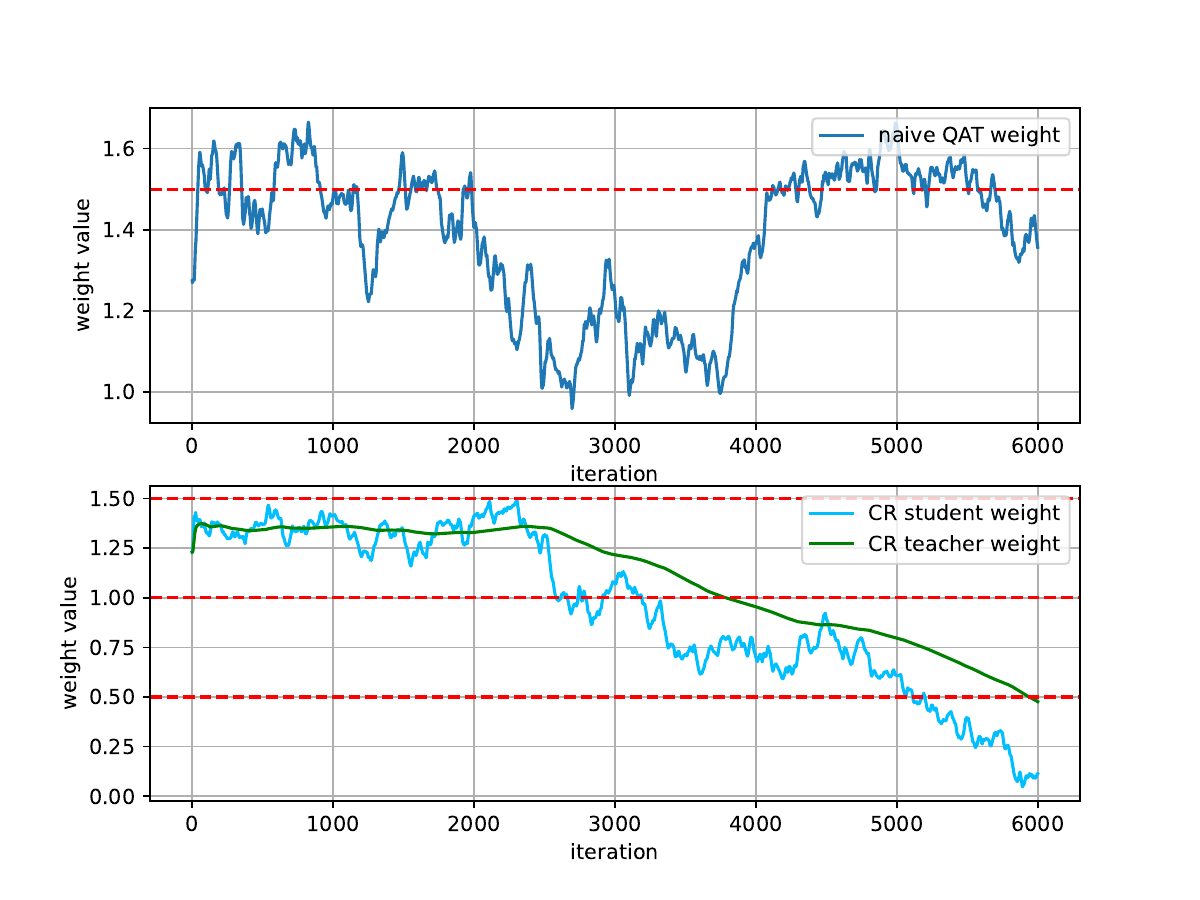}
    \caption{The weight training curve of the first channel in the first convolution of ResNet18 over 2000 iterations on CIFAR-10. The top row plots the weight for the naive QAT, while the bottom plots the weights for the student and teacher models in CR.}
    \label{fig:oscillations}
\end{figure}

\textbf{Generalization ability of CR:}~\cite{keskar-eigenvalue1-iclr-2017} and \cite{Wen-eigenvalue2-icais-2020} verified that the smaller the maximum eigenvalue of the Hessian matrix is, the flatter the landscape of loss is. Therefore, we compared the largest 50 eigenvalues of the FP32, LSQ, and our method for ResNet18 network in Fig.~\ref{fig:eigenvalues-comparision}. The results demonstrate that CR exhibited better generalization than that of the other methods.

\textbf{Entropy of CR:} The principle of maximum entropy indicates that a generalizable representation should be the one that admit the maximum entropy among all plausible representations. Consequently, we evaluated the entropy of the weights in neural networks. We built the histogram of weights on each kernel for each convolutional layer of the ResNet-18. The number of histogram bins was 70. The comparisons between LSQ and our method were shown in Tab.~\ref{tbl:entropy}.  Tab.~\ref{tbl:entropy} illustrated that the generalization ability of our method was better than that of the baseline model.

\begin{table}[t!]
\begin{center}
\centering
\caption{Comparison among FP, LSQ and our method with respect to information entropy. We utilize the ResNet-18 model and calculate the total sum of information entropy on all convolutional kernels within the model.}
\label{tbl:entropy}
\begin{tabular}{llll}
\toprule
Method              & {FP32}            & {LSQ}   &  {CR(Teacher)}       \\ \midrule
{Entropy} &{11803.78}        &{12570.15}   & {12599.46}             \\ \bottomrule                
\end{tabular}
\end{center}
\end{table}

\label{sec:weightoscillation}

\textbf{Handling Weight Oscillation Phenomenon} We visualized the value of a weight with respect to the number of iterations. Fig.~\ref{fig:oscillations} showed that the naive QAT exhibited severe weight oscillations around the quantization step boundaries; while CR effectively mitigated this problem. The explanation is that EMA would produce the smooth parameters, guiding the student model to learn the effective parameters.

\section{Conclusion}

In this paper, we find that low-bit quantized models has serious generalization problem. To address this problem, we propose CR, a simple, novel, yet compelling paradigm, for QAT. CR performs augmented an input image into two distinctive samples, where the reasonable perturbation is injected into the input. We theoretically that optimizing CR is equivalent to make the loss landscape flat. The flatter of loss landscape empirically brings more generalization ability to a neural network. Extensive experiments indicate that our method successfully enhances the generalization ability of the quantized.

Our empirical analysis is based on the classification tasks. How to design the CR loss for regression tasks, generation tasks, \textit{etc.}, is also worth exploring. Besides, how to keep the reasonable Lipschitz constant is also an important problem. 

\bibliographystyle{IEEEtran}
\bibliography{cite}

\end{document}